\documentclass[11pt,a4paper]{article}
\usepackage{times,latexsym}
\usepackage{url}
\usepackage[T1]{fontenc}

\usepackage{longtable}
\usepackage{setspace}
\usepackage{lipsum}
\usepackage{etoolbox}
\usepackage{multicol}
\usepackage{listings}
\usepackage{fancyvrb}
\usepackage{color}
\usepackage{pifont}
\usepackage[utf8]{inputenc} 
\usepackage[T1]{fontenc}    
\usepackage[T1]{fontenc}    
\usepackage{url}            
\usepackage{booktabs}       
\usepackage{amsfonts}       
\usepackage{amsmath}
\usepackage{nicefrac}       
\usepackage{microtype}      
\usepackage{multirow}
\usepackage{caption}
\usepackage{subcaption}
\usepackage{xcolor}
\usepackage{float}
\usepackage{xspace}
\usepackage{multirow}
\usepackage{graphicx}
\usepackage[linesnumbered,ruled,vlined]{algorithm2e}
\usepackage{amssymb}
\usepackage{colortbl}
\usepackage{adjustbox}
\usepackage{longtable} 

\usepackage[acceptedWithA]{tacl2021v1}
\usepackage{xspace,mfirstuc,tabulary}

\newif\iftaclinstructions
\taclinstructionsfalse 
\iftaclinstructions
\renewcommand{\confidential}{}
\renewcommand{\anonsubtext}{(No author info supplied here, for consistency with
TACL-submission anonymization requirements)}
\newcommand{\instr}
\fi

\iftaclpubformat 

\else

\fi

\title{Generative Induction of Dialogue Task Schemas with \\
Streaming Refinement and Simulated Interactions}

\author{
  James D. Finch
  \and
  Yasasvi Josyula
  \and
  Jinho D. Choi
  \\
  Department of Computer Science
  \\
  Emory University
  \\
  Atlanta, GA, USA
  \\
  \texttt{\{jdfinch, yasasvi.josyula, jinho.choi\}@emory.edu}
  \\
}

\date{}

\begin{document}
\maketitle
\begin{abstract}
  In task-oriented dialogue (TOD) systems, Slot Schema Induction (SSI) is essential for automatically identifying key information slots from dialogue data without manual intervention. This paper presents a novel state-of-the-art (SoTA) approach that formulates SSI as a text generation task, where a language model incrementally constructs and refines a slot schema over a stream of dialogue data. To develop this approach, we present a fully automatic LLM-based TOD simulation method that creates data with high-quality state labels for novel task domains. Furthermore, we identify issues in SSI evaluation due to data leakage and poor metric alignment with human judgment. We resolve these by creating new evaluation data using our simulation method with human guidance and correction, as well as designing improved evaluation metrics. These contributions establish a foundation for future SSI research and advance the SoTA in dialogue understanding and system development. 
\end{abstract}
\section{Introduction}
\label{sec:introduction}

Task-Oriented Dialogue (TOD) systems rely on a \textit{slot schema} to define the key types of information used to represent the state of the dialogue as it progresses towards task completion. Traditional approaches to slot schema creation require manual curation, which is both time-consuming and difficult to scale across domains \cite{rastogi_towards_2020}. Slot Schema Induction (SSI) has emerged as a solution, enabling automatic discovery of these slots from unlabeled dialogue data \cite{min_dialogue_2020}. This task plays a crucial role in advancing TOD research by reducing the need for manual schema creation, improving dialogue state tracking \cite{rana-etal-2025-zero}, and facilitating automated analysis of dialogue structure \cite{qiu_structure_2022}.

Despite the promise of SSI, current methods primarily rely on clustering dense embeddings of slot values, which may lose nuanced information about the relationship of each value to the overall dialogue \cite{finch-etal-2024-transforming, yu_unsupervised_2022}. Furthermore, these methods typically require large amounts of dialogue data up-front, which is not available for most dialogue systems until after their deployment. In this paper, we present a novel approach to SSI that breaks away from clustering slot-value embeddings by casting SSI as a text generation task that incrementally constructs and refines a slot schema over a stream of dialogue data. We develop generative models for our approach using both fine-tuning and prompting techniques, where the model is tasked to create new slots that capture key values discovered in the dialogue, while also tracking the values of previously-discovered slots to maintain a consistent schema. 

To develop this new method, we identify and address several critical challenges for training and evaluating SSI models. First, there is a lack of TOD data that covers a wide range of task domains, making it difficult to improve and evaluate the generalizability of SSI to novel settings. We address this by developing \textsc{Dots}, a fully automatic LLM-based simulation method to create TOD data with task schemas and state labels. Second, we conduct analyses that reveal flaws in existing SSI evaluation due to benchmark data leakage and poor agreement between performance metrics and human judgment. These are addressed by constructing new evaluation data using our \textsc{Dots} TOD simulation method with expert guidance and correction, and developing new evaluation metrics with superior agreement to human judgment. Using our new training data and evaluation setup, we conduct experiments comparing several variants of our method and previous approaches demonstrating that our approach achieves a new state of the art in SSI.


\section{Related Work}
\label{sec:related-work}

\paragraph{Dialogue Schema Induction}

TOD dialogue understanding involves creating structured representations to capture semantic information in dialogue text \cite{liang2024survey, budzianowski_multiwoz_2018}. Traditionally, these representations are manually defined for specific dialogue tasks, a labor-intensive process highlighting the need for automation \cite{agrawal2024dialog, liang2024survey}. Automating structured representation induction has been explored in various NLP tasks, including event schema induction \cite{dror-etal-2023-zero, mondal-etal-2025-adaptive}, text categorization \cite{an-etal-2024-generalized, liang-etal-2024-actively}, intent recognition \cite{liang-etal-2024-synergizing, rodriguez2024intentgpt}, and dialogue flow modeling \cite{el2023workflow, choubey2025turning}. Since our goal is towards inferring state representations for TOD, we focus on the task of Slot Schema Induction (SSI) in this work.

SSI conventionally involves two main steps: (1) extracting candidate values from dialogue text, and (2) clustering their semantic representations into slot types. Prior studies \cite{hudecek_discovering_2021, wu_semi-supervised_2022, qiu_structure_2022} use NLP tagging models for extraction, embedding candidates with an encoder such as BERT \cite{devlin-etal-2019-bert}, then clustering them using methods like similarity thresholding \cite{hudecek_discovering_2021} or iterative fine-tuning \cite{wu_semi-supervised_2022}. Some works use domain-general tools such as value taggers \cite{qiu_structure_2022}, PLM attention distributions \cite{yu_unsupervised_2022}, or slot-value generators \cite{finch-etal-2024-transforming} to discover new slot values. These methods all rely on clustering of dense value embedding vectors, which we hypothesize is suboptimal for representing slot semantics for SSI.
Thus we propose a novel text generation approach for SSI that does not rely on embedding-based value clustering.

\paragraph{Synthetic Data Generation}

An ongoing challenge in TOD research is the scarcity of high-quality data that spans a diverse set of task domains. MultiWOZ \cite{budzianowski_multiwoz_2018} and SGD \cite{rastogi_towards_2020} are currently the largest and most popular datasets by far. MultiWOZ collects human-human dialogue via a Wizard-of-Oz setup, while SGD uses rule-based dialogue simulation from handcrafted schemas with human paraphrasing. These datasets cover only 5 and 16 domains, respectively, due to their reliance on costly human annotation.

\noindent To overcome this limitation, prior work has explored data augmentation by synthesizing dialogues from human-provided task specifications, such as dialogue flows or schemas \cite{campagna_zero-shot_2020, aksu_velocidapter_2021, aksu_n-shot_2022, mehri_lad_2022, mohapatra_simulated_2021, kim_neuralwoz_2021, wan_unified_2022}. More recently, LLM-based methods have been proposed to fully automate dialogue generation, eliminating human involvement in both task specification and dialogue creation \cite{li2023camel, finch-choi-2024-diverse}. \citet{finch-choi-2024-diverse} further automate state annotation in the synthesized dialogues but produce inconsistent slot schemas and noisy annotations.
To the best of our knowledge, we are the first to achieve fully automated TOD simulation with a diverse set of consistent ground-truth slot schemas.

\paragraph{Evaluation Benchmark Leakage}
Concerns are mounting over the leakage of evaluation benchmarks to language models, such as OpenAI’s GPT \cite{balloccu-etal-2024-leak, golchintime, ranaldi-etal-2024-investigating, xu2024benchmark}. Notably for the purpose of evaluating slot schema induction approaches, \citet{balloccu-etal-2024-leak} revealed that the MultiWOZ 2.4 test set has been leaked to GPT. The discovery of data leakage complicates the evaluation of slot induction methods on existing datasets, as models have likely already seen the correct answers during training, making their performance unreliable as a true measure of generalization. In Section \ref{sec:evaluation-method-data}, we provide quantitative verification of MultiWOZ and SGD benchmark leakage to popular LLMs and introduce a new, unexposed evaluation dataset to overcome this challenge.

\definecolor{lightblue}{rgb}{0.678, 0.847, 0.902}   
\definecolor{lightgreen}{rgb}{0.678, 0.902, 0.678}  
\definecolor{lightyellow}{rgb}{1.0, 1.0, 0.678}     
\definecolor{lightorange}{rgb}{1.0, 0.831, 0.547}   
\definecolor{lightpink}{rgb}{1.0, 0.761, 0.796}     
\definecolor{lightpurple}{rgb}{0.768, 0.631, 0.902}  
\definecolor{lightcyan}{rgb}{0.678, 1.0, 1.0}       
\definecolor{lightgray}{rgb}{0.847, 0.847, 0.847}    

\renewcommand{\arraystretch}{0.8}
\begin{figure}[htb!]
    \centering    
    \adjustbox{max width=\columnwidth}{%
    \begin{tabular}{|p{1.25\columnwidth}|}
        \hline
        \rowcolor{lightblue} \\
        \rowcolor{lightblue} \textbf{\# Key Information Types} \\
        \rowcolor{lightblue} \\
        \rowcolor{lightblue} \#\# Garden Layouts \\
        \rowcolor{lightblue} * \textit{style:} The preferred style of the garden layout. \\
        \rowcolor{lightblue} * \textit{features:} Special features included in the layout. \\
        \rowcolor{lightblue} * \textit{maintenance\_level:} The level of maintenance required. \\
        \rowcolor{lightblue} \\
        \rowcolor{lightblue} \#\# Plant Selections \\
        \rowcolor{lightblue} * \textit{type:} The type of plant, such as Flower, Shrub, Tree, or Grass. \\
        \rowcolor{lightblue} * \textit{color:} The color preference for the plant's blooms or foliage. \\
        \rowcolor{yellow!30} \\
        \rowcolor{yellow!30} \textbf{\# Dialogue} \\
        \rowcolor{yellow!30} \textit{Gardener:} I'm looking for a small desert garden layout with a \\ \rowcolor{yellow!30} \quad fountain; can you help? \\
        \rowcolor{yellow!30} ... \\
        \rowcolor{yellow!30} \textit{Gardener:} Could you help me find some pink flowers that thrive \\
        \rowcolor{yellow!30} \quad in full sun for my garden? \\
        \rowcolor{yellow!30} \textit{Landscaper:} Sure! How about the Desert Rose or Ice Plant? \\
        \rowcolor{yellow!30} \quad They both have pink flowers and thrive in full sun! \\
        \rowcolor{yellow!30} \textit{Gardener:} I love the sound of the Desert Rose! Can you tell me \\
        \rowcolor{yellow!30} \quad about its water requirements? \\
        \rowcolor{yellow!30} \textit{Landscaper:} The Desert Rose has medium water requirements, \\
        \rowcolor{yellow!30} \quad so it needs a bit more moisture than some other desert plants. \\
        \rowcolor{yellow!30} \textit{Gardener:} I see, would it be possible to find a pink flower with \\
        \rowcolor{yellow!30} \quad low water requirements instead? \\
        \rowcolor{orange!30} \\
        \rowcolor{orange!30} Identify Key Information Values from the Dialogue \\
        \toprule
        \bottomrule
        \rowcolor{green!30}  \\
        \rowcolor{green!30} \# \textbf{Key Information Values} \\
        \rowcolor{green!30} \\

        \begin{tabular}{@{}p{0.63\columnwidth} @{}p{0.62\columnwidth}}  
        \rowcolor{green!30} \#\# Garden Layouts & \#\# Plant Selections \\
        \rowcolor{green!30} * \textit{style:} desert & * \textit{type:} Flower \\
        \rowcolor{green!30} * \textit{features:} fountain & * \textit{color:} Pink \\
        \rowcolor{green!30} * \textit{maintenance\_level:} low & * \textit{sunlight:} Full Sun \\
        \rowcolor{green!30} \textit{  } & - the plant's sun requirements \\
        \end{tabular} \\
        \hline
    \end{tabular}
    }
    \caption{Example token sequence from \textsc{Dots} (\textsection \ref{sec:dialogue-simulation}) used to train SSI model $M$. [\texttt{YELLOW}]: dialogue context $D$, [\texttt{BLUE}]: slot schema $\mathbb{S}_{t-1}$, [\texttt{ORANGE}]: instruction, [\texttt{GREEN}]: predicted state $S_t$.}
    \label{fig:dsi-sequence-format}
    \vspace{-2ex}
\end{figure}

\section{Approach}
\label{sec:approach}

This work proposes using language model text generation to incrementally construct and refine a slot schema from a stream of dialogue data. Given a partial slot schema $\mathbb{S}_{t-1}$, and the last dialogue $D$ in the data stream, our approach infers an updated schema $\mathbb{S}_t$ and dialogue state $S_t$ for turn $T_t$. By initializing an empty schema $\mathbb{S}_0 = \emptyset$, this approach can be used for SSI by iterating over any dialogue datset, with the final schema $\mathbb{S}_{final}$ serving as the overall schema prediction.

Each updated schema $\mathbb{S}_t$ and state $S_t$ are inferred jointly as a single generated text sequence using a language model $M$. An example sequence from our training data is shown in Figure \ref{fig:dsi-sequence-format} to exemplify the format. As shown, our approach uses a joint formulation of zero-shot DST and new slot discovery, expanding upon prior works where these tasks were addressed separately
\cite{gupta_show_2022, finch-etal-2024-transforming, dong-etal-2024-zero}. 
Jointly tackling DST and new slot discovery in a single generated sequence is key: by conditioning generation on previous schema $\mathbb{S}_{t-1}$, the objective of model $M$ is to predict the values of \textit{existing slots} whenever possible, ensuring minimal redundancy in the slot schema across various dialogues. However, the objective of $M$ also predicts when there are important values in $D$ that lack existing slots. In these cases, $M$ generates new slots with descriptive natural language labels, expanding the schema coverage to appropriately capture the underlying task structure. This process also facilitates the discovery of new task domains, since all slots are identified with a task domain name $d$ in addition to a slot name $s$. Given the generated text prediction at turn $t$, the state $S_t$ is simply the set of domain-slot-value triples $(d, s, v)$ parsed from the text, and the updated schema $\mathbb{S}_t$ is calculated as:
\vspace{-1ex}
\begin{equation*}
    \mathbb{S}_t = \mathbb{S}_{t-1} \cup \{(d,s):(d,s,v)\in S_t\}
\end{equation*}

\noindent While this approach is sufficient for achieving streaming schema induction, it may produce noisy slot discoveries. Without a revision or filtering mechanism, the induced schema grows monotonically as dialogues are processed, eventually leading to degenerate inferences or implementation issues when the schema length exceeds the model's handling capacity. To address this challenge, we investigate two mechanisms for revising the schema by removing or modifying a subset of its slots: Schema Revision (\textsection \ref{sec:approach-schema-revision-noisy-schemas}) and Slot Confidence Estimation (\textsection \ref{sec:approach-slot-confidence-window-filter}).

\subsection{Schema Revision}
\label{sec:approach-schema-revision-noisy-schemas}

To revise noisy slot inferences, we train language model $M$ to additionally revise noisy schema $\mathbb{S}_t^n$ predicted by an SSI model $N$ into ground-truth schema $\mathbb{S}_t$. This uses a sequence format similar to Figure \ref{fig:dsi-sequence-format}, but where the target prediction sequence represents only $\mathbb{S}_t$ without state information $S_t$. However, a challenge to this revision training is that any SSI model $N$ used will likely predict an $\mathbb{S}_t^n$ with an entirely different surface form compared to $\mathbb{S}_t$, even for slots with similar meaning to the ground truth (for example, naming a slot "max price" versus "budget"). Therefore, training $M$ to predict $\mathbb{S}_t$ by directly revising $\mathbb{S}_t^n$ causes a model that rewrites the schema entirely every turn, causing a high degree of variance in the inferred schema. To improve the stability of revision prediction, we focus instead on \textit{adding noise} to ground-truth $\mathbb{S}_t$ using $\mathbb{S}_t^n$ to obtain the schema to be revised $\mathbb{S}_t^*$, which trains $M$ to revise the subset of $\mathbb{S}_t^*$ that is predicted to be noise. We focus on three strategies of adding noise, randomly selecting between them for each training sequence:
\vspace{-0.5ex}
\begin{enumerate}
    \item No noise is added, training $M$ to detect when the schema needs no revision.
    \item A random subset of $\mathbb{S}_t^n$ is added to $\mathbb{S}_t$, training $M$ to revise duplicate slots.
    \item A random subset of $\mathbb{S}_t^n$ is combined with a random subset of $\mathbb{S}_t$, training $M$ to rewrite, deduplicate, or add missing slots.
\end{enumerate}




\begin{figure*}[htb!]
    \centering
    \includegraphics[width=\textwidth]{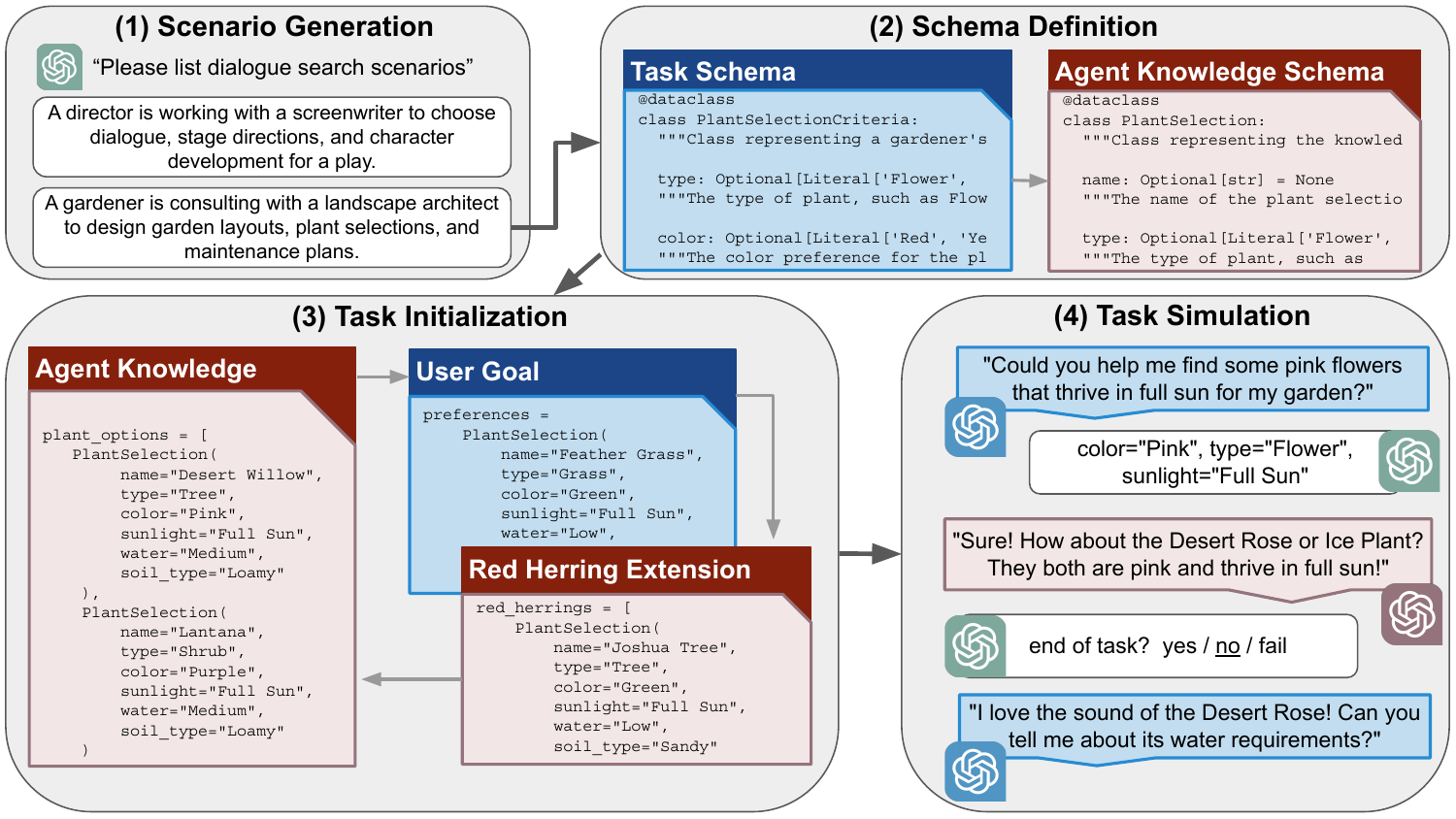}
    \caption{Overview of the dialogue simulation used to create the \textsc{Dots} dataset.}
    \label{fig:dialogue-simulation}
    \vspace{-2ex}
\end{figure*}

\subsection{Slot Confidence Estimation}
\label{sec:approach-slot-confidence-window-filter}

In addition to modeling schema revision as another text generation subtask of our SSI approach, we observe that simply measuring the frequency with which slots in the inferred schema $\hat{\mathbb{S}}$ are assigned values in state inferences $\hat{S}$ can be used as a confidence measure of slots' suitability for representing the dialogue state. By this principle, slots that are rarely updated with values in tracked states $\hat{S}$ are likely noise and should be removed from the schema. We investigate this hypothesis by developing an alternative approach to Schema Revision. 

Given a window size $w$, each slot in the induced schema $s \in \hat{\mathbb{S}}_t$ is removed from the schema if in recent states $\hat{S}_{t-w..t}$ it was assigned a value $c < \tau$ times, where $\tau$ represents a confidence threshold and $c$ is the number of times its value was updated. This simple approach has the limitation of including important density-based parameters, similar to previous work's reliance on HDBSCAN clustering \cite{finch-etal-2024-transforming, yu_unsupervised_2022}; however, we demonstrate in our experiments (\textsection \ref{sec:results}) that it reliably filters out noisy slot discoveries.

\section{Task-Oriented Dialogue Simulation}
\label{sec:dialogue-simulation}

Given the limited task coverage of existing TOD training and evaluation data, we investigate a TOD simulation method inspired by \citet{rastogi_towards_2020} and \citet{finch-choi-2024-diverse} to automatically generate diverse dialogues with high quality schemas and state annotations. Summarized in Figure \ref{fig:dialogue-simulation}, the method is broken down into four parts: Scenario Generation, Schema Definition, Task Initialization, and Task Simulation. Each of these parts is performed automatically using an LLM with zero-shot instruction. Inspired by \citet{king_diverse_2023}, many of our prompts instruct the LLM to write a python script with a dataclass definition or object instantiation, which we find to be a reliable method for generating structured data such as slot schemas and dialogue state labels. We use GPT-4o and GPT-4o-mini for all simulation steps.\footnote{gpt-4o-2024-08-06 and gpt-4o-mini-2024-07-18}

\paragraph{Scenario Generation} Multi-domain scenarios are automatically generated as a numbered list of templated one-sentence descriptions in the form: \textit{"<user> is getting help from <agent> in order to <A>, <B>, ..."}. Each listed sentence represents a scenario $\sigma$ with speaker tags for user $U$ and agent $A$ roles, and an ordered list of task domains $T$.

\paragraph{Schema Definition} For each task $t\in T$ in a scenario $\sigma$, the structure of the task is defined by generating a slot schema $\mathbb{S}$ and agent knowledge schema $\mathbb{K}$. The slot schema $\mathbb{S}$ is generated first as a python dataclass based on the scenario description $\sigma$ and task label $t$, where fields in the dataclass each represent a type of preference or requirement that user $U$ might bring to the task. The agent knowledge schema $\mathbb{K}$ is similarly generated based on slot schema $\mathbb{S}$, $t$, and $\sigma$, to represent the structure of actual knowledge held by agent $A$. For example, if $\mathbb{S}$ has a slot called "max price" representing a preference, $\mathbb{K}$ should have a corresponding field like "price" to represent an actual value.

\paragraph{Task Initialization} To begin simulating each TOD dialogue for a scenario $\sigma$, the schemas of its first task $t = T_1$ are used to initialize the knowledge $K$ of agent $A$ and the goal $G$ of user $U$. $K$ is generated as a list of python objects constructed using the dataclass representation $\mathbb{K}$, and is meant to serve as a repository of candidate solutions to $t$. From this list, one item is randomly selected to serve as an "ideal" solution $K_i$. The task goal $G$ is then generated to represent the preferences and requirements of user $U$ based on the ideal $K_i$ by instantiating task schema $\mathbb{S}$. A random subset of slots in $G$ are cleared to represent no preference. To increase the difficulty of the task, a list of red herrings $H$ is then created to extend $K$ with items similar to $G$. Finally, ideal solution $K_i$ may be removed from $K$ a random 50\% of the time to cover the realistic situation of no optimal solution. In these cases, the user $U$ may need to change or compromise their preferences to finish the task.

\paragraph{Task Simulation} After initializing a task $t$, alternating dialogue generation representing utterances for $U$ and $A$ builds the dialogue $D$. Because dialogue generation for $U$ is conditioned only on $D$ and $G$, while generation for $A$ only sees $D$ and $K$, the two simulated agents must converse to share goal and knowledge information, making a non-trivial and fairly realistic interaction. The dialogue generation prompt encourages short responses to avoid all relevant information being shared on the first turn. After each user turn, the current dialogue state is tracked given $D$ and $\mathbb{S}$ by instructing the LLM to instantiate the schema dataclass to represent all preference information shared by $U$. The task simulation is detected by prompted end-of-task classification that checks $D$ after every agent turn. When the end of the task is detected, the dialogue either continues by initializing the next task in $T$ for scenario $\sigma$, or ends if no tasks remain.

\subsection{\textsc{Dots} Training Data}
\label{sec:dialogue-simulation-dots-training-data}

Using the presented \textsc{Dots} simulation method, we generate a TOD dataset in order to train a slot discovery model that uses the approach described in Section \ref{sec:approach}. The training data is created with $N=300$ scenarios, 18 of which are manually removed due to domain overlap with either MultiWOZ or the 10 scenarios of the \textsc{Dots} test set described in the next section (\textsection \ref{sec:evaluation-method-data}). Given the generated scenario descriptions and schemas for the remaining 282 scenarios, the Task Initialization and Simulation steps generate 10 dialogues per scenario. 49 dialogues are lost due to syntax issues or other errors in generated output. The final 2,771 simulated dialogues are summarized and compared to other TOD training data in Table \ref{tab:train-data-stats}.

\begin{table}[htbp]
    \centering
    \resizebox{\columnwidth}{!}{
    \begin{tabular}{@{}l c c c c c c@{}}
        \toprule
        \bf Data & \bf Dial.s & \bf Turns & \bf Dom.s & \bf Slots & \bf Values \\
        \midrule
        \textsc{Mwoz} & 8,420 & 104,916 & 5 & 31 & 56,668 \\
        \textsc{Sgd}  & 16,142 & 329,964 & 16 & 115 & 164,982 \\
        \textsc{Dot}  & 5,015 & 100,471 & 1,003 & 173,572* & 487,460 \\
        \midrule
        \textsc{Dots} & 2,771 & 88,240 & 787$^\dagger$ & 6,810$^\dagger$ & 44,120 \\
        \bottomrule
    \end{tabular}
    }
    \caption{Training split statistics of MultiWOZ, SGD, \textsc{Dot} \cite{finch-choi-2024-diverse}, and \textsc{Dots}. * slots are not semantically unique. $\dagger$ slots are unique per domain, but domains sometimes overlap.}
    \label{tab:train-data-stats}
    \vspace{-2ex}
\end{table}

\section{Evaluation}
\label{sec:evaluation-method}

This work addresses two critical issues for current SSI benchmark evaluation. First, we observe that TOD data has leaked into the training of the most popular base models. This issue is presented and addressed in Section \ref{sec:evaluation-method-data} with a new test set for SSI, created using the \textsc{Dots} TOD simulation pipeline with manual refinement and correction. Second, the metrics used to evaluate the previous SoTA SSI approaches heavily overestimate performance and disagree strongly with human judgment. Section \ref{sec:evaluation-method-metric} addresses this by presenting a validation analysis of the metrics used in previous work and a new metric with vastly superior agreement to human judgment.

\subsection{New TOD Test Data}
\label{sec:evaluation-method-data}

During pilot experiments, GPT-4o, Claude 3.5-Sonnet, and Llama 3.1 and 3.2 variants were observed to be capable of predicting the slot schemas of SGD and MultiWOZ \textit{without any dialogue data present in their context}, merely an instruction to recall the slot schema given the name of the dataset. To quantify benchmark data leakage into these models' training data with respect to SSI, we conduct a human evaluation where one of the authors prompts various base models to recall the slot schema of either MultiWOZ or SGD from their parametric knowledge. Table \ref{tab:data-leakage-analysis} reports the Precision, Recall, and F1 scores of the slots recalled by each model compared to the gold schema, revealing that all tested models are compromised. 

\begin{table}[htbp]
    \centering
    \resizebox{\columnwidth}{!}{
    \begin{tabular}{@{}l ccc ccc@{}}
        \toprule
         & \multicolumn{3}{c}{MultiWOZ} & \multicolumn{3}{c}{SGD} \\
        \cmidrule(lr){2-4} \cmidrule(lr){5-7}
        \bf Model & \bf P & \bf R & \bf F1 & \bf P & \bf R & \bf F1 \\
        \midrule
        Llama-3B & 25.0 & 16.7  & 20.0  & 35.6 & \multicolumn{1}{r}{9.5}  & 15.0 \\
        Llama-8B & 45.2 & 46.7  & 45.9  & 71.2  & 21.9 & 33.5 \\
        Claude-3.5S & 100.0 & 100.0 & 100.0 & 74.3 & 59.8  & 66.2 \\
        GPT-4o & 100.0 & 100.0 & 100.0 & 84.2 & 50.3 & 63.0 \\
        \bottomrule
    \end{tabular}
    }
    \caption{Ability of various models to recall the slot schemas of MultiWOZ and SGD from their parametric knowledge alone when instructed, providing evidence of benchmark data leakage for SSI.}
    \label{tab:data-leakage-analysis}
    \vspace{-2ex}
\end{table}

Given that no SSI benchmark data are available whose schemas have not been leaked into the training of relevant base models, a new evaluation dataset is constructed to fill this gap. Ideally, the test data would be collected using recordings of a wide range of real-world naturalistic dialogue scenarios with expert-annotated task schemas and dialogue states. However, such data collection is extraordinarily expensive and difficult to achieve. Thus, the methods used to create previous benchmark data MultiWOZ and SGD were semi-automatic. The creation of our new data broadly follows SGD's approach, but takes advantage of LLM's high-quality dialogue generation ability using our \textsc{Dots} simulation pipeline:

\paragraph{Test Domains} are created using 10 handcrafted scenario descriptions written by one of the authors. 5 scenarios include 2 domains and 5 include 3, making 25 unique task domains in total. Using a handcrafted approach, the scenarios are guaranteed to reflect plausible TOD dialogue applications that are also unique among each other and compared to domains in MultiWOZ and SGD.

\paragraph{Domain Schemas} are created by feeding handcrafted scenario descriptions to our Schema Definition pipeline. To ensure the resulting schemas are a valid reflection of the task scenario, all 25 domain task schemas are manually corrected by 2 of the authors. In general, the schemas are high-quality: from 212 original slots, 4 new slot types were created to address potential constraints for successful task completion that were missing, 8 nonsensical slot types were removed, and 2 slots were revised to provide additional clarity in their name and description, resulting in a final total of 208 slots across the 25 domains.

\paragraph{State-Annotated Dialogues} are created by feeding the corrected schemas into the Task Initialization and Task Simulation stages of the simulation pipeline. 100 dialogues were simulated per scenario. From these 1,000 dialogues, 30 per scenario were then manually cherry-picked and then corrected. Hand-selecting a subset of 300 dialogues from a larger set of 1,000 allows avoiding degenerate dialogues\footnote{In rare instances, simulated dialogues stall on a task or end prematurely if the end-of-task classification step fails.} and improves the diversity of task goals and dialogue progression within the same scenarios. Among the final set of 300 dialogues, 7.2\% of turns were revised to improve naturalness, and 4.3\% of slot-value pairs were corrected due to errors produced by the automatic state annotator.

\begin{table}[htbp]
    \centering
    \resizebox{\columnwidth}{!}{
    \begin{tabular}{@{}l c c c c c c@{}}
        \toprule
        \bf Data & \bf Dial.s & \bf Turns & \bf Dom.s & \bf Slots & \bf Values \\
        \midrule
        \textsc{Mwoz} & 999 & 13,737 & 5 & 30 & 7,368 \\
        \textsc{Sgd} & 4,201 & 84,594 & 18 & 106 & 42,297 \\
        \midrule
        \textsc{Dots} & 300 & 7,844 & 25 & 208 & 3,922 \\
        \bottomrule
    \end{tabular}
    }
    \caption{Statistics of SSI test data.}
    \label{tab:test-data-stats}
    \vspace{-1ex}
\end{table}

\noindent Although the \textsc{Dots} test set is smaller at a little more than half the size of MultiWOZ, it has the highest degree of domain diversity among the three evaluation data compared in Table \ref{tab:test-data-stats} with almost twice as many slot types as SGD. This diversity makes it more suitable as a test set for SSI, since approaches are tested on their ability to induce schemas across a wide range of diverse dialogue tasks. In fact, because \textsc{Dots} includes 10 unique and disjoint multi-domain application scenarios, it facilitates estimating the \textit{expected performance} for schema induction on any novel scenario by averaging performance across independent per-scenario evaluations. This allows more realistic and reliable estimation of SSI performance than was possible with previous evaluation data MultiWOZ and SGD, which arguably each only represent a single application scenario. Our experiments thus follow this paradigm of estimating expected performance by averaging per-scenario evaluation metrics. 

\subsection{Improved Evaluation Metrics}
\label{sec:evaluation-method-metric}

Evaluation of SSI aims to measure the quality of an induced set of slots $P$ by matching it against a gold reference slot schema $G$ \cite{finch-etal-2024-transforming, yu_unsupervised_2022}. Each predicted slot \( p \in P \) and gold slot \( g \in G \) are represented as sets of context-value pairs \( (c, v) \), where \( c \) is a particular dialogue context and \( v \) is a value filling the slot. Given predicted slots and gold slots $P$ and $G$, an SSI evaluator defines a mapping \( M: P \to G \) that associates each predicted slot \( p \in P \) to the reference slot \( g \in G \) that has the highest similarity score \( S(p, g) \):

\begingroup
\setlength{\abovedisplayskip}{1pt}
\setlength{\belowdisplayskip}{5pt}
\[
M^* = \{ (p, g) \mid p\in P,  g = \arg\max_{g \in G} S(p, g) \}
\]
\endgroup

\noindent Previous SoTA SSI approaches are evaluated automatically \cite{finch-etal-2024-transforming, yu_unsupervised_2022}, where $S$ is implemented as $S_{BERT}$ by computing the centroid of each induced and gold reference slot cluster using BERT encodings \cite{devlin-etal-2019-bert} of their values. Each induced cluster $p$ is mapped to the gold slot $g$ whose cluster centroid is nearest by cosine similarity, or not mapped if there is no match of 80\% similarity or higher. See \citet{finch-etal-2024-transforming} for further detail.

We find this automatic matching method based on embedding similarity is extremely noisy and does not reflect human judgments of the slot mapping. Thus, we propose the alternative similarity function \( S_{exact}(p, g) \) calculated as the precision of values in $p$ that exactly match (caseless) those in $g$ and use an adjusted similarity threshold of $x < 0.5$ for determining no match:

\vspace{-1ex}
\[
    S_{\text{exact}}(p, g) = \frac{|p \cap g|}{|p|}
\]
\vspace{-3ex}

\paragraph{Validation Study} To validate each automatic matcher, we perform a validation experiment using a human-judged slot mapping $M_h$. Using the previous SoTA SSI method from \citet{finch-etal-2024-transforming} to infer predicted slots $P$ on the \textsc{Dots} evaluation data, $M_h$ was constructed by one of the authors over the 170 predicted slots. The human mapping $M_h$ is used as a ground truth reference to validate automatically created mappings using the BERT-based similarity function $S_{BERT}$ from previous work and the proposed similarity function $S_{exact}$. 

Only \textbf{35.3\%} of predicted slots were correctly mapped using $S_{BERT}$, compared to \textbf{71.8\%} correctly mapped slots using $S_{exact}$. Disagreements between $M_h$ and $\hat M$ predicted using $S_{exact}$ were, as expected, usually due to overly strict similarity estimation producing no match in cases where the human found a match, or due to impure slot clusters in $P$ combining the semantics of multiple slots in $G$, which creates ambiguous mapping decisions.


\paragraph{Metrics} Table \ref{tab:evaluation_metrics} presents the formulas for the evaluation metrics used in our experiments, which are calculated given mapping $\hat{M}$ constructed by our improved automatic matcher.
Note that, unlike the presented metric equations, previous work allows multiple predicted slots to be mapped to the same gold slot \cite{finch-etal-2024-transforming, yu_unsupervised_2022}. This results in vastly inflated scores when calculating precision, and encourages approaches that focus on optimizing recall by predicting an order of magnitude more slots than what appear in gold schemas. Our metrics adjust the precision calculation to count each redundant slot prediction as a precision error.

\begin{table}[ht]
    \centering
    \resizebox{\columnwidth}{!}{
    \renewcommand{\arraystretch}{1.1} 
    \begin{tabular}{l|cc}
        \hline
        & \textbf{Precision} & \textbf{Recall} \\
        \hline
        \\[-8pt]
        \textbf{Slot} & 
        $\displaystyle\frac{|\{g:(p,g)\in M\}|}{|P|}$ & 
        $\displaystyle\frac{|\{g:(p,g)\in M\}|}{|G|}$ \\[14pt]
        
        \textbf{Value} & 
        $\displaystyle\frac{\sum_{(p,g) \in M} |p \cap g|}{\sum_{(p,g) \in M} |p|}$ & 
        $\displaystyle\frac{\sum_{(p,g) \in M} |p \cap g|}{\sum_{(p,g) \in M} |g|}$ \\[14pt]
        \hline
    \end{tabular}
    }
    \caption{Precision and Recall formulas for evaluation metrics of induced slots and discovered values.}
    \label{tab:evaluation_metrics}
    \vspace{-2.5ex}
\end{table}

\section{Experiments}
\label{sec:experiments}

Experiments use the \textsc{Dots} evaluation data and refined metrics from Section \ref{sec:evaluation-method}. Since slot induction aims to induce a schema for tracking dialogue state in novel settings, we assess performance across multi-domain scenarios by evaluating slot induction on each of the 10 test scenarios independently and reporting the average for each metric.\footnote{All experiment results are also averaged over 3 replicates to address any stochastic variance like dialogue stream order.} We assess the performance of our proposed methods against previous works and strong baselines. Each included method is a unique combination of approach (\textsection \ref{sec:exp-approach}), state representation (\textsection \ref{sec:exp-state}), training data (\textsection \ref{sec:exp-training}), and base model (\textsection \ref{sec:exp-models}).

\subsection{Approaches}
\label{sec:exp-approach}

\paragraph{\textsc{Revision}} is our main streaming slot induction approach using the revision mechanism (\textsection \ref{sec:approach-schema-revision-noisy-schemas}). Revision training data is obtained from the outputs of a \textsc{Slot Conf} model trained on a small development version of \textsc{Dots} separate from the training and evaluation splits, thus producing noisy outputs that satisfy the revision setup. 

\paragraph{\textsc{Slot Conf}} is our main streaming slot induction approach using slot update counts to estimate confidence in each slot for filtering noisy discoveries (\textsection \ref{sec:approach-slot-confidence-window-filter}). The window length is set to 10 dialogues and the confidence threshold to 1 update.

\paragraph{\textsc{Fifo}} is a baseline simplifying \textsc{Slot Conf}, which filters out the least-recently-filled slots when the predicted schema size exceeds 100 slots.

\paragraph{\textsc{Priority}} is another baseline of \textsc{Slot Conf} that maintains global slot update counts, removing the least frequently updated slots when the schema reaches 100 slots.

\paragraph{\textsc{Embed}} follows prior slot induction work from \citet{finch-etal-2024-transforming} where SBERT embeddings of induced slots are clustered to predict slot schemas. \textsc{Embed} models trained on \textsc{Dot} embed slot name + value, while models trained on \textsc{Dots} perform better when embedding predicted slot descriptions. Clustering uses HDBSCAN \cite{mcinnes_hdbscan_2017} with hyperparameters selected from a grid search optimizing silhouette score \cite{yu_unsupervised_2022}.

\vspace{1.5ex}
\noindent Note that, because our streaming slot induction approach (\textsection \ref{sec:approach}) incrementally updates schema predictions, all experiments with streaming methods use a two-pass setup. In pass 1, the final schema is determined, and then pass 2 runs in "DST mode," ignoring new slot discoveries, to allow the model to represent earlier states in the dialogue data with the final slot schema. This ensures a fairer comparison to clustering methods without penalizing streaming approaches for early schema revisions.

\subsection{State Representations}
\label{sec:exp-state}

The main approach presented in Section \ref{sec:approach} is described to predict dialogue state representations for the most recent dialogue turn, which is similar to dialogue state trackers \cite{king_diverse_2023, gupta_show_2022, finch-choi-2024-diverse}. However, the approach can be easily modified to predict only the \textit{updates} to the dialogue state made by the most recent dialogue turn, similar to previous slot induction work \cite{finch-etal-2024-transforming, yu_unsupervised_2022}, or to make slot induction predictions on the last turn of the dialogue only.

\paragraph{\textsc{State}} predicts full dialogue states including values from the entire dialogue history, similar to DST work, for all turns in each dialogue.

\paragraph{\textsc{Update}} predicts only dialogue state updates from the most recent turn, following prior slot induction work.

\paragraph{\textsc{Final}} modifies \textsc{State} to predict states only at the end of each dialogue, which may reduce noisy new slot discoveries caused by a lack of context for how the dialogue progresses.

\subsection{Training Data}
\label{sec:exp-training}

\paragraph{\textsc{Dots}} is the new data presented in Section \ref{sec:dialogue-simulation-dots-training-data}.

\paragraph{\textsc{Dot}} \cite{finch-choi-2024-diverse} lacks ground-truth schemas, preventing training of streaming approaches, but is used to train variants of \textsc{Embed}.

\paragraph{\textsc{Sgdx}} augments SGD \cite{rastogi_towards_2020}, the most domain-diverse existing dataset with ground-truth schemas, by replacing each training sequences' slots and descriptions with one of their six SGD-X \cite{lee_sgd-x_2022} variants.

\subsection{Base Models}
\label{sec:exp-models}

\paragraph{\textsc{Llama-8B/3B/1B}} Instruct 3.1/3.2 variants \cite{dubey2024llama} are fine-tuned using QLoRA with rank 1 and alpha 2, applying adapters to all linear layers, following findings that rank has little impact when applied consistently across all linear layers \cite{dettmers_qlora_2023}. Training uses the Adam optimizer with a learning rate of $10^{-4}$, batch size of 8, and 30,000 steps.

\paragraph{\textsc{T5-3B}} \cite{raffel2020exploring} is evaluated for direct comparison with previous work, using the publicly released model from \citet{finch-etal-2024-transforming} that was trained on \textsc{Dot}.

\paragraph{\textsc{Claude}} 3.5-Sonnet \cite{anthropic2024claude} is assessed. As a closed-access model, it is evaluated via API with zero-shot instruction for slot induction instead of fine-tuning.

\section{Results}
\label{sec:results}

\paragraph{Comparison of Approaches} Table~\ref{tab:dsi-results} presents a comparison of the streaming and embedding approaches trained on \textsc{Dots} using various state representations. Of the three main approach categories, the \textsc{Slot Conf} based streaming methods appear to perform the best, with \textsc{Slot Conf} predicting \textsc{Final} states achieving both the highest Slot F1 (66.8) and Value F1 (74.9). The \textsc{Revision} streaming methods performed second-best, although they are outperformed by even the baseline \textsc{Slot Conf} approaches: \textsc{Fifo} and \textsc{Priority}. Furthermore, the \textsc{Embed} approach from previous work performed worst, which confirms our hypothesis that using embedding clustering for slot induction is unsuitable,  although the \textsc{Final} state variant achieved high Slot Precision (74.3) with high-quality clusters at a Value F1 of 74.8. In general, prediction of \textsc{Final} states appears more reliable than \textsc{State} or \textsc{Update} alternatives, across all methodologies.

\begin{table}[htbp]
    \centering
    \resizebox{\columnwidth}{!}{
    \begin{tabular}{@{}l c ccccc c}
        \toprule
         & & \multicolumn{3}{c}{\bf Slot} & \multicolumn{3}{c}{\bf Value} \\
        \cmidrule(lr){3-5} \cmidrule(lr){6-8}
        \bf Approach & \bf S & \bf P & \bf R & \bf F1 & \bf P & \bf R & \bf F1 \\
        \midrule
         \textsc{Embed} & U & 39.9 & 37.6 & 25.5 & 73.3 & 29.9 & 38.0 \\
          & F & \textbf{74.3} & 39.4 & 46.2 & 76.3 & \textbf{77.4} & \underline{74.8} \\
          \midrule
         \textsc{Fifo} & F & 55.2 & 72.1 & \underline{62.0} & 81.5 & 64.7 & 71.7 \\
         \textsc{Priority} & F & 54.0 & 70.8 & 60.6 & 81.4 & 63.8 & 71.2 \\
         \textsc{Slot Conf} & U & 50.9 & \underline{73.1} & 59.4 & \underline{84.1} & 45.4 & 57.8 \\
          & S & 45.9 & 70.5 & 55.0 & 82.7 & 54.7 & 66.5 \\
          & F & \underline{62.5} & 72.6 & \textbf{66.8} & 82.5 & \underline{69.2} & \textbf{74.9}  \\
          \midrule
         \textsc{Revision} & U & 52.3 & 68.3 & 58.2 & \textbf{85.6} & 44.3 & 56.9 \\
          & S & 43.3 & \textbf{73.5} & 53.2 & 80.4 & 50.9 & 61.4  \\
          & F & 50.4 & 70.5 & 57.8 & 80.5 & 60.0 & 68.2 \\
        \bottomrule
    \end{tabular}
    }
    \caption{Average Precision/Recall/F1 (\textbf{P}/\textbf{R}/\textbf{F1}) on induced slots and discovered values for Llama-8B models trained for different combinations of induction approaches and state representations (\textbf{S:} U (\textsc{Update}), S (\textsc{State}), F (\textsc{Final})). \textbf{Highest} scores are bolded and \underline{second-highest} are underlined.}
    \label{tab:dsi-results}
    \vspace{-1ex}
\end{table}

\paragraph{Comparison of Domain-Diverse Data} We can directly assess the relative utility of the \textsc{Dots} dataset against the similarly domain-diverse existing dataset \textsc{Dot} by comparing models using the \textsc{Embed} methodology but trained on the two different datasets. The results in Table \ref{tab:dot-results} confirm that training on the \textsc{Dots} data substantially enhance schema induction performance compared to \textsc{Dot}, especially in the \textsc{Final} state setup which is only possible for the schema-consistent \textsc{Dots} data. This is evidence that the structured and simulated \textsc{Dots} generation method reduces noise and improves reliability of both slot induction and value discovery.

\begin{table}[htbp]
    \centering
    \resizebox{\columnwidth}{!}{
    \begin{tabular}{@{}l c c ccccc c}
        \toprule
         & & & \multicolumn{3}{c}{\bf Slot} & \multicolumn{3}{c}{\bf Value} \\
        \cmidrule(lr){4-6} \cmidrule(lr){7-9}
        \bf Data & \bf M & \bf S & \bf P & \bf R & \bf F1 & \bf P & \bf R & \bf F1 \\
        \midrule
          \textsc{Dot} & T5 & U & 22.9 & \bf 39.5 & 21.1 & 63.3 & 20.3 & 27.6 \\
          \textsc{Dot} & Ll & U & 45.3 & 20.3 & 22.0 & 73.4 & 45.5 & 53.6 \\
          \midrule
          \textsc{Dots} & Ll & U & 39.9 & 37.6 & 25.5 & 73.3 & 29.9 & 38.0 \\
          \textsc{Dots} & Ll & F & \bf 74.3 & 39.4 & \bf 46.2 & \bf 76.3 & \bf 77.4 & \bf 74.8 \\
        \bottomrule
    \end{tabular}
    }
    \caption{Average Precision/Recall/F1 (\textbf{P}/\textbf{R}/\textbf{F1}) on induced slots and discovered values for Llama-8B \textsc{Embed} models (\textbf{Ll}) trained on \textsc{Dot} or \textsc{Dots} against \textsc{Embed} approach using \textsc{Dot}-trained T5 model from \citet{finch-etal-2024-transforming}.}
    \label{tab:dot-results}
    \vspace{-2ex}
\end{table}

\paragraph{Comparison of Schema-Consistent Data} When comparing against the existing schema-consistent dataset \textsc{SGD}, we find \textsc{Dots} to be the superior training resource for the novel streaming slot induction method (Table \ref{tab:sgd-results}). For both \textsc{Slot Conf} and \textsc{Revision} approaches, \textsc{Dots} trained models achieve better performance, likely reflecting the utility of the increased diversity of training data.

\begin{table}[htbp]
    \centering
    \resizebox{\columnwidth}{!}{
    \begin{tabular}{@{}l c ccccc c}
        \toprule
         & & \multicolumn{3}{c}{\bf Slot} & \multicolumn{3}{c}{\bf Value} \\
        \cmidrule(lr){3-5} \cmidrule(lr){6-8}
        \bf Data & \bf Appr. & \bf P & \bf R & \bf F1 & \bf P & \bf R & \bf F1 \\
        \midrule
         \textsc{Sgdx} & \textsc{S.C.} & 47.5 & 43.8 & 44.7 & 75.1 & 50.5 & 59.6 \\
         \textsc{Sgdx} & \textsc{Rev.} & 51.9 & 19.1 & 27.5 & 71.4 & 55.5 & 61.5 \\
         \midrule
         \textsc{Dots} & \textsc{S.C.} & \bf 62.5 & \bf 72.6 & \bf 66.8 & \bf 82.5 & \bf 69.2 & \bf 74.9  \\
         \textsc{Dots} & \textsc{Rev.} & 50.4 & 70.5 & 57.8 & 80.5 & 60.0 & 68.2 \\
        \bottomrule
    \end{tabular}
    }
    \caption{Average Precision/Recall/F1 (\textbf{P}/\textbf{R}/\textbf{F1}) on induced slots and discovered values for Llama-8B \textsc{Slot Conf} models (\textbf{S.C.}) or \textsc{Revision} models (\textbf{Rev.}) trained on \textsc{SGDX} or \textsc{Dots} using \textsc{Final} state representations.}
    \label{tab:sgd-results}
    \vspace{-1ex}
\end{table}

\begin{figure}[ht]
    \centering
    \includegraphics[width=0.95\columnwidth]{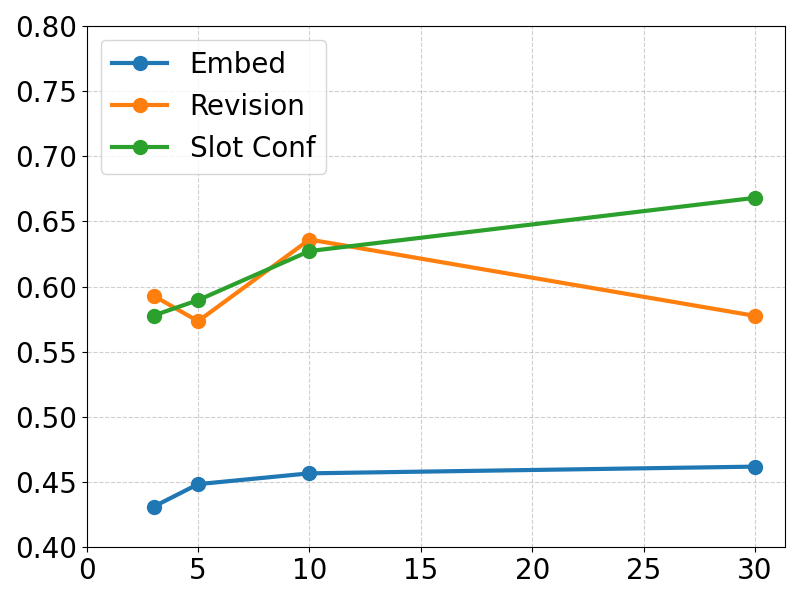}
    \caption{Slot F1 Score vs. Number of Dialogues}
    \label{fig:lines}
    \vspace{-1ex}
\end{figure}

\paragraph{Low Resource Slot Discovery} Figure~\ref{fig:lines} illustrates the relationship between the number of dialogues and slot F1 score for our best-performing \textsc{Slot Conf}, \textsc{Revision}, and \textsc{Embed} models. The results suggest that \textsc{Slot Conf} schema induction benefits the most from increased data, while \textsc{Embed} remains relatively stagnant, and \textsc{Revision} exhibits inconsistent performance trends.

\paragraph{Model Size Comparisons} Table \ref{tab:model-results} compares the performance of different model sizes trained using our best approach, \textsc{Slot Conf} with \textsc{Final} state inference, against \textsc{Claude}. Among all methods evaluated, \textsc{Claude} in \textsc{Slot Conf} mode achieves the highest overall performance, with an average Slot F1 of 69.2 and an average Value F1 of 84.3. Our fine-tuned \textsc{Llama-8B} model performs competitively, coming within 2.5 points of \textsc{Claude} in Slot F1 and 11 points in Value F1. As expected, smaller Llama models show a decline in performance, with the 3B and 1B models trailing by 8 points and 11 points in Slot F1.

\begin{table}[htbp]
    \centering
    \resizebox{\columnwidth}{!}{
    \begin{tabular}{@{}l c ccccc c}
        \toprule
         & & \multicolumn{3}{c}{\bf Slot} & \multicolumn{3}{c}{\bf Value} \\
        \cmidrule(lr){3-5} \cmidrule(lr){6-8}
        \bf Model & \bf Appr. & \bf P & \bf R & \bf F1 & \bf P & \bf R & \bf F1 \\
        \midrule
         \textsc{Ll-1B} & \textsc{S.C.} & 32.4 & 49.2 & 38.6 & 71.6 & 49.4 & 57.2 \\
         \textsc{Ll-3B} & \textsc{S.C.} & 54.5 & 64.8 & 58.5 & 77.9 & 64.1 & 69.7 \\
         \textsc{Ll-8B} & \textsc{S.C.} & 62.5 & \bf 72.6 & 66.8 & \bf 82.5 & 69.2 & 74.9 \\
         \textsc{Ll-8B} & \textsc{Rev.} & 50.4 & 70.5 & 57.8 & 80.5 & 60.0 & 68.2 \\
         \midrule
         \textsc{Claude} & \textsc{S.C.} & \bf 74.2 & 65.7 & \bf 69.2 & 80.8 & 88.9 & 84.3 \\
         \textsc{Claude} & \textsc{Rev.} & 64.3 & 56.5 & 59.7 & 81.1 & \bf 93.5 & \bf 86.5 \\
        \bottomrule
    \end{tabular}
    }
    \caption{Average Precision/Recall/F1 (\textbf{P}/\textbf{R}/\textbf{F1}) on induced slots and discovered values for Llama (\textbf{Ll}) \textsc{Slot Conf} models (\textbf{S.C.}) of various sizes trained on \textsc{Dots} with \textsc{Final} state representation, against Llama \textsc{Revision} model (\textbf{Rev.}) and Claude \textsc{Slot Conf} and \textsc{Revision} models using the same state.}
    \label{tab:model-results}
    \vspace{-1ex}
\end{table}

To further investigate this comparison, we perform a human evaluation of the predicted slot schemas from our best \textsc{Llama-8B} model and \textsc{Claude} using the method described in Section \ref{sec:evaluation-method-metric}. Both approaches benefit from less strict matching between predicted and gold slots, achieving 81.8 and 84.6 Slot F1 respectively (Table \ref{tab:human-model-results}).

\begin{table}[htbp]
    \centering
    \resizebox{0.6\columnwidth}{!}{
    \begin{tabular}{@{}l ccc}
        \toprule
         & \multicolumn{3}{c}{\bf Slot} \\
        \cmidrule(lr){2-4}
        \bf Model & \bf P & \bf R & \bf F1 \\
        \midrule
         \textsc{Ll-8B} & 76.8 & \bf 88.2 & 81.8 \\
         \textsc{Claude} & \bf 91.3 & 80.0 & \bf 84.6 \\
        \bottomrule
    \end{tabular}
    }
    \caption{Average Precision/Recall/F1 (\textbf{P}/\textbf{R}/\textbf{F1}) on induced slots based on human mapping of predicted-to-gold slots for best-performing models: Llama-8B \textsc{Slot Conf} (\textbf{Ll-8B}) and Claude \textsc{Slot Conf} (\textbf{Claude}).}
    \label{tab:human-model-results}
    \vspace{-1ex}
\end{table}

\section{Discussion}  
\label{sec:discussion}

\paragraph{Induction as Text Generation} is found to be a powerful approach for slot induction. Our experiments highlight the advantages of jointly inferring newly discovered slots while tracking existing ones through streaming generation, significantly outperforming embedding-based clustering. Furthermore, integrating DST into this approach provides a simple yet effective confidence measure for each discovered slot, enabling the filtering of noisy slots as more dialogue data becomes available. 
Surprisingly, the text generation–based schema refinement approach, \textsc{Revision}, does not improve upon the other studied refinement method, \textsc{Slot Conf}, or its baselines, \textsc{Fifo} and \textsc{Priority}. Manual analysis suggests the limitation of \textsc{Revision} to be that it produces high variance schema inferences. This may be caused by overly local revision decisions made based on single dialogue examples, whereas the simpler \textsc{Slot Conf} method benefits from confidence estimations across many dialogues. One possible direction for future work is to explore LLM-based refinement strategies that incorporate global statistical patterns, perhaps by encoding such information as part of the model's input, in efforts to further improve and explore the capabilities of this joint approach to SSI.

\paragraph{TOD Data Creation} remains an ongoing research challenge, but our work demonstrates the power of LLM-based simulation for constructing TOD resources for training and evaluation. The \textsc{Dots} automatic data generation method is the first to create TOD data for novel diverse tasks with ground-truth task schemas. The strategy of generating dialogues as simulations from generated task schemas turns out to produce higher quality state-labeled dialogue data than the noisier automatic-annotation strategy from previous work \cite{finch-choi-2024-diverse}, and allows for multi-domain and schema-consistent dialogue creation at low cost. TOD simulation methods may continue to alleviate the difficult and costly nature of collecting data, and future work may help to further improve the realism and applicability of our presented methods.

\paragraph{Slot Induction Evaluation} benchmarks from previous work are invalidated due to the data leakage and evaluation metric issues demonstrated in our analyses (\textsection \ref{sec:evaluation-method}). We address these issues by providing new validated evaluation metrics and evaluation data created with human guidance and correction. However, there may be room for further improvement, since the automatically-generated nature of the \textsc{Dots} evaluation data inevitably carries biases compared to the data distribution of real-world TOD settings. Although costly, investment in creating more realistic and diverse TOD datasets may be a key step forward for all TOD research, especially since other subtasks such as DST may be affected by data leakage as well.

\section{Conclusion}

The presented work marks a large step forward for SSI, advancing TOD research. Our findings highlight the importance of text generation-based approaches for slot induction, the value of simulation-based data creation, and the necessity of rigorous evaluation methodologies for advancing TOD research. We publicly release the new SoTA slot induction models and \textsc{Dots} data to facilitate future work. These contributions improve automatic TOD understanding and support dialogue system development, helping to create more natural and effective human-computer interactions.

\bibliography{tacl2021, anthology}

\begin{thebibliography}{42}
\expandafter\ifx\csname natexlab\endcsname\relax\def\natexlab#1{#1}\fi

\bibitem[{Agrawal et~al.(2024)Agrawal, Pillai, Uppuluri, Reddy, Li, Tur, Hakkani-Tur, and Ji}]{agrawal2024dialog}
Stuti Agrawal, Pranav Pillai, Nishi Uppuluri, Revanth~Gangi Reddy, Sha Li, Gokhan Tur, Dilek Hakkani-Tur, and Heng Ji. 2024.
\newblock Dialog flow induction for constrainable llm-based chatbots.
\newblock In \emph{SIGDIAL}.

\bibitem[{Aksu et~al.(2022)Aksu, Liu, Kan, and Chen}]{aksu_n-shot_2022}
Ibrahim Aksu, Zhengyuan Liu, Min-Yen Kan, and Nancy Chen. 2022.
\newblock \href {https://doi.org/10.18653/v1/2022.findings-acl.131} {N-{Shot} {Learning} for {Augmenting} {Task}-{Oriented} {Dialogue} {State} {Tracking}}.
\newblock In \emph{Findings of the {Association} for {Computational} {Linguistics}: {ACL} 2022}, pages 1659--1671, Dublin, Ireland. Association for Computational Linguistics.

\bibitem[{Aksu et~al.(2021)Aksu, Liu, Kan, and Chen}]{aksu_velocidapter_2021}
Ibrahim~Taha Aksu, Zhengyuan Liu, Min-Yen Kan, and Nancy Chen. 2021.
\newblock \href {https://aclanthology.org/2021.sigdial-1.14} {Velocidapter: {Task}-oriented {Dialogue} {Comprehension} {Modeling} {Pairing} {Synthetic} {Text} {Generation} with {Domain} {Adaptation}}.
\newblock In \emph{Proceedings of the 22nd {Annual} {Meeting} of the {Special} {Interest} {Group} on {Discourse} and {Dialogue}}, pages 133--143, Singapore and Online. Association for Computational Linguistics.

\bibitem[{An et~al.(2024)An, Shi, Tian, Lin, Wang, Wu, Cai, Wang, Chen, Zhu, and Chen}]{an-etal-2024-generalized}
Wenbin An, Wenkai Shi, Feng Tian, Haonan Lin, QianYing Wang, Yaqiang Wu, Mingxiang Cai, Luyan Wang, Yan Chen, Haiping Zhu, and Ping Chen. 2024.
\newblock \href {https://doi.org/10.18653/v1/2024.findings-acl.512} {Generalized category discovery with large language models in the loop}.
\newblock In \emph{Findings of the Association for Computational Linguistics: ACL 2024}, pages 8653--8665, Bangkok, Thailand. Association for Computational Linguistics.

\bibitem[{Anthropic(2024)}]{anthropic2024claude}
AI~Anthropic. 2024.
\newblock Claude 3.5 sonnet model card addendum.
\newblock \emph{Claude-3.5 Model Card}, 3(6).

\bibitem[{Balloccu et~al.(2024)Balloccu, Schmidtov{\'a}, Lango, and Dusek}]{balloccu-etal-2024-leak}
Simone Balloccu, Patr{\'i}cia Schmidtov{\'a}, Mateusz Lango, and Ondrej Dusek. 2024.
\newblock \href {https://aclanthology.org/2024.eacl-long.5/} {Leak, cheat, repeat: Data contamination and evaluation malpractices in closed-source {LLM}s}.
\newblock In \emph{Proceedings of the 18th Conference of the European Chapter of the Association for Computational Linguistics (Volume 1: Long Papers)}, pages 67--93, St. Julian{'}s, Malta. Association for Computational Linguistics.

\bibitem[{Budzianowski et~al.(2018)Budzianowski, Wen, Tseng, Casanueva, Ultes, Ramadan, and Gašić}]{budzianowski_multiwoz_2018}
Pawe{\textbackslash}l Budzianowski, Tsung-Hsien Wen, Bo-Hsiang Tseng, Iñigo Casanueva, Stefan Ultes, Osman Ramadan, and Milica Gašić. 2018.
\newblock \href {https://doi.org/10.18653/v1/D18-1547} {{MultiWOZ} - {A} {Large}-{Scale} {Multi}-{Domain} {Wizard}-of-{Oz} {Dataset} for {Task}-{Oriented} {Dialogue} {Modelling}}.
\newblock In \emph{Proceedings of the 2018 {Conference} on {Empirical} {Methods} in {Natural} {Language} {Processing}}, pages 5016--5026, Brussels, Belgium. Association for Computational Linguistics.

\bibitem[{Campagna et~al.(2020)Campagna, Foryciarz, Moradshahi, and Lam}]{campagna_zero-shot_2020}
Giovanni Campagna, Agata Foryciarz, Mehrad Moradshahi, and Monica Lam. 2020.
\newblock \href {https://doi.org/10.18653/v1/2020.acl-main.12} {Zero-{Shot} {Transfer} {Learning} with {Synthesized} {Data} for {Multi}-{Domain} {Dialogue} {State} {Tracking}}.
\newblock In \emph{Proceedings of the 58th {Annual} {Meeting} of the {Association} for {Computational} {Linguistics}}, pages 122--132, Online. Association for Computational Linguistics.

\bibitem[{Choubey et~al.(2025)Choubey, Peng, Bhagavath, Xiong, Pentyala, and Wu}]{choubey2025turning}
Prafulla~Kumar Choubey, Xiangyu Peng, Shilpa Bhagavath, Caiming Xiong, Shiva~Kumar Pentyala, and Chien-Sheng Wu. 2025.
\newblock Turning conversations into workflows: A framework to extract and evaluate dialog workflows for service ai agents.
\newblock \emph{arXiv preprint arXiv:2502.17321}.

\bibitem[{Dettmers et~al.(2023)Dettmers, Pagnoni, Holtzman, and Zettlemoyer}]{dettmers_qlora_2023}
Tim Dettmers, Artidoro Pagnoni, Ari Holtzman, and Luke Zettlemoyer. 2023.
\newblock \href {https://proceedings.neurips.cc/paper_files/paper/2023/hash/1feb87871436031bdc0f2beaa62a049b-Abstract-Conference.html} {{QLoRA}: {Efficient} {Finetuning} of {Quantized} {LLMs}}.
\newblock \emph{Advances in Neural Information Processing Systems}, 36:10088--10115.

\bibitem[{Devlin et~al.(2019)Devlin, Chang, Lee, and Toutanova}]{devlin-etal-2019-bert}
Jacob Devlin, Ming-Wei Chang, Kenton Lee, and Kristina Toutanova. 2019.
\newblock \href {https://doi.org/10.18653/v1/N19-1423} {{BERT}: Pre-training of deep bidirectional transformers for language understanding}.
\newblock In \emph{Proceedings of the 2019 Conference of the North {A}merican Chapter of the Association for Computational Linguistics: Human Language Technologies, Volume 1 (Long and Short Papers)}, pages 4171--4186, Minneapolis, Minnesota. Association for Computational Linguistics.

\bibitem[{Dong et~al.(2024)Dong, Feng, Lu, Shi, and Wu}]{dong-etal-2024-zero}
Xiaoyu Dong, Yujie Feng, Zexin Lu, Guangyuan Shi, and Xiao-Ming Wu. 2024.
\newblock \href {https://doi.org/10.18653/v1/2024.emnlp-main.485} {Zero-shot cross-domain dialogue state tracking via context-aware auto-prompting and instruction-following contrastive decoding}.
\newblock In \emph{Proceedings of the 2024 Conference on Empirical Methods in Natural Language Processing}, pages 8527--8540, Miami, Florida, USA. Association for Computational Linguistics.

\bibitem[{Dror et~al.(2023)Dror, Wang, and Roth}]{dror-etal-2023-zero}
Rotem Dror, Haoyu Wang, and Dan Roth. 2023.
\newblock \href {https://doi.org/10.18653/v1/2023.findings-eacl.53} {Zero-shot on-the-fly event schema induction}.
\newblock In \emph{Findings of the Association for Computational Linguistics: EACL 2023}, pages 705--725, Dubrovnik, Croatia. Association for Computational Linguistics.

\bibitem[{Dubey et~al.(2024)Dubey, Jauhri, Pandey, Kadian, Al-Dahle, Letman, Mathur, Schelten, Yang, Fan et~al.}]{dubey2024llama}
Abhimanyu Dubey, Abhinav Jauhri, Abhinav Pandey, Abhishek Kadian, Ahmad Al-Dahle, Aiesha Letman, Akhil Mathur, Alan Schelten, Amy Yang, Angela Fan, et~al. 2024.
\newblock The llama 3 herd of models.
\newblock \emph{arXiv preprint arXiv:2407.21783}.

\bibitem[{El~Hattami et~al.(2023)El~Hattami, Laradji, Raimondo, V{\'a}zquez, Rodr{\'\i}guez, and Pal}]{el2023workflow}
Amine El~Hattami, Issam~H Laradji, Stefania Raimondo, David V{\'a}zquez, Pau Rodr{\'\i}guez, and Christopher Pal. 2023.
\newblock Workflow discovery from dialogues in the low data regime.
\newblock \emph{Trans. Mach. Learn. Res.}

\bibitem[{Finch and Choi(2024)}]{finch-choi-2024-diverse}
James~D. Finch and Jinho~D. Choi. 2024.
\newblock \href {https://doi.org/10.18653/v1/2024.findings-emnlp.731} {Diverse and effective synthetic data generation for adaptable zero-shot dialogue state tracking}.
\newblock In \emph{Findings of the Association for Computational Linguistics: EMNLP 2024}, pages 12527--12544, Miami, Florida, USA. Association for Computational Linguistics.

\bibitem[{Finch et~al.(2024)Finch, Zhao, and Choi}]{finch-etal-2024-transforming}
James~D. Finch, Boxin Zhao, and Jinho~D. Choi. 2024.
\newblock \href {https://doi.org/10.18653/v1/2024.sigdial-1.27} {Transforming slot schema induction with generative dialogue state inference}.
\newblock In \emph{Proceedings of the 25th Annual Meeting of the Special Interest Group on Discourse and Dialogue}, pages 317--324, Kyoto, Japan. Association for Computational Linguistics.

\bibitem[{Golchin and Surdeanu(2024)}]{golchintime}
Shahriar Golchin and Mihai Surdeanu. 2024.
\newblock Time travel in llms: Tracing data contamination in large language models.
\newblock In \emph{The Twelfth International Conference on Learning Representations}.

\bibitem[{Gupta et~al.(2022)Gupta, Lee, Zhao, Cao, Rastogi, and Wu}]{gupta_show_2022}
Raghav Gupta, Harrison Lee, Jeffrey Zhao, Yuan Cao, Abhinav Rastogi, and Yonghui Wu. 2022.
\newblock \href {https://doi.org/10.18653/v1/2022.naacl-main.336} {Show, {Don}'t {Tell}: {Demonstrations} {Outperform} {Descriptions} for {Schema}-{Guided} {Task}-{Oriented} {Dialogue}}.
\newblock In \emph{Proceedings of the 2022 {Conference} of the {North} {American} {Chapter} of the {Association} for {Computational} {Linguistics}: {Human} {Language} {Technologies}}, pages 4541--4549, Seattle, United States. Association for Computational Linguistics.

\bibitem[{Hudeček et~al.(2021)Hudeček, Dušek, and Yu}]{hudecek_discovering_2021}
Vojtěch Hudeček, Ondřej Dušek, and Zhou Yu. 2021.
\newblock \href {https://doi.org/10.18653/v1/2021.acl-long.189} {Discovering {Dialogue} {Slots} with {Weak} {Supervision}}.
\newblock In \emph{Proceedings of the 59th {Annual} {Meeting} of the {Association} for {Computational} {Linguistics} and the 11th {International} {Joint} {Conference} on {Natural} {Language} {Processing} ({Volume} 1: {Long} {Papers})}, pages 2430--2442, Online. Association for Computational Linguistics.

\bibitem[{Kim et~al.(2021)Kim, Chang, and Lee}]{kim_neuralwoz_2021}
Sungdong Kim, Minsuk Chang, and Sang-Woo Lee. 2021.
\newblock \href {https://doi.org/10.18653/v1/2021.acl-long.287} {{NeuralWOZ}: {Learning} to {Collect} {Task}-{Oriented} {Dialogue} via {Model}-{Based} {Simulation}}.
\newblock In \emph{Proceedings of the 59th {Annual} {Meeting} of the {Association} for {Computational} {Linguistics} and the 11th {International} {Joint} {Conference} on {Natural} {Language} {Processing} ({Volume} 1: {Long} {Papers})}, pages 3704--3717, Online. Association for Computational Linguistics.

\bibitem[{King and Flanigan(2023)}]{king_diverse_2023}
Brendan King and Jeffrey Flanigan. 2023.
\newblock \href {https://doi.org/10.18653/v1/2023.findings-acl.344} {Diverse {Retrieval}-{Augmented} {In}-{Context} {Learning} for {Dialogue} {State} {Tracking}}.
\newblock In \emph{Findings of the {Association} for {Computational} {Linguistics}: {ACL} 2023}, pages 5570--5585, Toronto, Canada. Association for Computational Linguistics.

\bibitem[{Lee et~al.(2022)Lee, Gupta, Rastogi, Cao, Zhang, and Wu}]{lee_sgd-x_2022}
Harrison Lee, Raghav Gupta, Abhinav Rastogi, Yuan Cao, Bin Zhang, and Yonghui Wu. 2022.
\newblock \href {https://doi.org/10.1609/aaai.v36i10.21341} {{SGD}-{X}: {A} {Benchmark} for {Robust} {Generalization} in {Schema}-{Guided} {Dialogue} {Systems}}.
\newblock \emph{Proceedings of the AAAI Conference on Artificial Intelligence}, 36(10):10938--10946.
\newblock Number: 10.

\bibitem[{Li et~al.(2023)Li, Hammoud, Itani, Khizbullin, and Ghanem}]{li2023camel}
Guohao Li, Hasan Hammoud, Hani Itani, Dmitrii Khizbullin, and Bernard Ghanem. 2023.
\newblock Camel: Communicative agents for" mind" exploration of large language model society.
\newblock \emph{Advances in Neural Information Processing Systems}, 36:51991--52008.

\bibitem[{Liang et~al.(2024{\natexlab{a}})Liang, Liao, Fei, and Jiang}]{liang-etal-2024-synergizing}
Jinggui Liang, Lizi Liao, Hao Fei, and Jing Jiang. 2024{\natexlab{a}}.
\newblock \href {https://doi.org/10.18653/v1/2024.findings-acl.840} {Synergizing large language models and pre-trained smaller models for conversational intent discovery}.
\newblock In \emph{Findings of the Association for Computational Linguistics: ACL 2024}, pages 14133--14147, Bangkok, Thailand. Association for Computational Linguistics.

\bibitem[{Liang et~al.(2024{\natexlab{b}})Liang, Liao, Fei, Li, and Jiang}]{liang-etal-2024-actively}
Jinggui Liang, Lizi Liao, Hao Fei, Bobo Li, and Jing Jiang. 2024{\natexlab{b}}.
\newblock \href {https://doi.org/10.18653/v1/2024.naacl-long.434} {Actively learn from {LLM}s with uncertainty propagation for generalized category discovery}.
\newblock In \emph{Proceedings of the 2024 Conference of the North American Chapter of the Association for Computational Linguistics: Human Language Technologies (Volume 1: Long Papers)}, pages 7845--7858, Mexico City, Mexico. Association for Computational Linguistics.

\bibitem[{Liang et~al.(2024{\natexlab{c}})Liang, Wu, Fang, Fei, and Liao}]{liang2024survey}
Jinggui Liang, Yuxia Wu, Yuan Fang, Hao Fei, and Lizi Liao. 2024{\natexlab{c}}.
\newblock A survey of ontology expansion for conversational understanding.
\newblock In \emph{Proceedings of the 2024 Conference on Empirical Methods in Natural Language Processing}, pages 18111--18127.

\bibitem[{McInnes et~al.(2017)McInnes, Healy, and Astels}]{mcinnes_hdbscan_2017}
Leland McInnes, John Healy, and Steve Astels. 2017.
\newblock \href {https://doi.org/10.21105/joss.00205} {hdbscan: {Hierarchical} density based clustering}.
\newblock \emph{Journal of Open Source Software}, 2(11):205.

\bibitem[{Mehri et~al.(2022)Mehri, Altun, and Eskenazi}]{mehri_lad_2022}
Shikib Mehri, Yasemin Altun, and Maxine Eskenazi. 2022.
\newblock \href {https://aclanthology.org/2022.sigdial-1.55} {{LAD}: {Language} {Models} as {Data} for {Zero}-{Shot} {Dialog}}.
\newblock In \emph{Proceedings of the 23rd {Annual} {Meeting} of the {Special} {Interest} {Group} on {Discourse} and {Dialogue}}, pages 595--604, Edinburgh, UK. Association for Computational Linguistics.

\bibitem[{Min et~al.(2020)Min, Qin, Teng, Liu, and Zhang}]{min_dialogue_2020}
Qingkai Min, Libo Qin, Zhiyang Teng, Xiao Liu, and Yue Zhang. 2020.
\newblock Dialogue state induction using neural latent variable models.
\newblock In \emph{Proceedings of the {Twenty}-{Ninth} {International} {Joint} {Conference} on {Artificial} {Intelligence}}, {IJCAI}'20, pages 3845--3852, Yokohama, Yokohama, Japan.

\bibitem[{Mohapatra et~al.(2021)Mohapatra, Pandey, Contractor, and Joshi}]{mohapatra_simulated_2021}
Biswesh Mohapatra, Gaurav Pandey, Danish Contractor, and Sachindra Joshi. 2021.
\newblock \href {https://doi.org/10.18653/v1/2021.findings-emnlp.103} {Simulated {Chats} for {Building} {Dialog} {Systems}: {Learning} to {Generate} {Conversations} from {Instructions}}.
\newblock In \emph{Findings of the {Association} for {Computational} {Linguistics}: {EMNLP} 2021}, pages 1190--1203, Punta Cana, Dominican Republic. Association for Computational Linguistics.

\bibitem[{Mondal et~al.(2025)Mondal, Yuan, N, Garimella, Ferraro, Blair-Stanek, Van~Durme, and Boyd-Graber}]{mondal-etal-2025-adaptive}
Ishani Mondal, Michelle Yuan, Anandhavelu N, Aparna Garimella, Francis Ferraro, Andrew Blair-Stanek, Benjamin Van~Durme, and Jordan Boyd-Graber. 2025.
\newblock \href {https://aclanthology.org/2025.coling-main.392/} {{ADAPTIVE} {IE}: Investigating the complementarity of human-{AI} collaboration to adaptively extract information on-the-fly}.
\newblock In \emph{Proceedings of the 31st International Conference on Computational Linguistics}, pages 5870--5889, Abu Dhabi, UAE. Association for Computational Linguistics.

\bibitem[{Qiu et~al.(2022)Qiu, Wu, Liu, and Xiong}]{qiu_structure_2022}
Liang Qiu, Chien-Sheng Wu, Wenhao Liu, and Caiming Xiong. 2022.
\newblock \href {https://doi.org/10.48550/arXiv.2203.00073} {Structure {Extraction} in {Task}-{Oriented} {Dialogues} with {Slot} {Clustering}}.
\newblock ArXiv:2203.00073 [cs].

\bibitem[{Raffel et~al.(2020)Raffel, Shazeer, Roberts, Lee, Narang, Matena, Zhou, Li, and Liu}]{raffel2020exploring}
Colin Raffel, Noam Shazeer, Adam Roberts, Katherine Lee, Sharan Narang, Michael Matena, Yanqi Zhou, Wei Li, and Peter~J Liu. 2020.
\newblock Exploring the limits of transfer learning with a unified text-to-text transformer.
\newblock \emph{Journal of machine learning research}, 21(140):1--67.

\bibitem[{Rana et~al.(2025)Rana, Hacioglu, Gopalan, and Boothalingam}]{rana-etal-2025-zero}
Mansi Rana, Kadri Hacioglu, Sindhuja Gopalan, and Maragathamani Boothalingam. 2025.
\newblock \href {https://aclanthology.org/2025.coling-industry.59/} {Zero-shot slot filling in the age of {LLM}s for dialogue systems}.
\newblock In \emph{Proceedings of the 31st International Conference on Computational Linguistics: Industry Track}, pages 697--706, Abu Dhabi, UAE. Association for Computational Linguistics.

\bibitem[{Ranaldi et~al.(2024)Ranaldi, Ruzzetti, Onorati, Ranaldi, Giannone, Favalli, Romagnoli, and Zanzotto}]{ranaldi-etal-2024-investigating}
Federico Ranaldi, Elena~Sofia Ruzzetti, Dario Onorati, Leonardo Ranaldi, Cristina Giannone, Andrea Favalli, Raniero Romagnoli, and Fabio~Massimo Zanzotto. 2024.
\newblock \href {https://doi.org/10.18653/v1/2024.findings-acl.827} {Investigating the impact of data contamination of large language models in text-to-{SQL} translation}.
\newblock In \emph{Findings of the Association for Computational Linguistics: ACL 2024}, pages 13909--13920, Bangkok, Thailand. Association for Computational Linguistics.

\bibitem[{Rastogi et~al.(2020)Rastogi, Zang, Sunkara, Gupta, and Khaitan}]{rastogi_towards_2020}
Abhinav Rastogi, Xiaoxue Zang, Srinivas Sunkara, Raghav Gupta, and Pranav Khaitan. 2020.
\newblock \href {https://doi.org/10.1609/aaai.v34i05.6394} {Towards {Scalable} {Multi}-{Domain} {Conversational} {Agents}: {The} {Schema}-{Guided} {Dialogue} {Dataset}}.
\newblock \emph{Proceedings of the AAAI Conference on Artificial Intelligence}, 34(05):8689--8696.
\newblock Number: 05.

\bibitem[{Rodriguez et~al.(2024)Rodriguez, Botzer, Vazquez, Pal, Pedersoli, and Laradji}]{rodriguez2024intentgpt}
Juan~A Rodriguez, Nicholas Botzer, David Vazquez, Christopher Pal, Marco Pedersoli, and Issam~H Laradji. 2024.
\newblock Intentgpt: Few-shot intent discovery with large language models.
\newblock In \emph{ICLR 2024 Workshop on Large Language Model (LLM) Agents}.

\bibitem[{Wan et~al.(2022)Wan, Zhang, Zhu, Liao, and Huang}]{wan_unified_2022}
Dazhen Wan, Zheng Zhang, Qi~Zhu, Lizi Liao, and Minlie Huang. 2022.
\newblock \href {https://doi.org/10.18653/v1/2022.findings-emnlp.277} {A {Unified} {Dialogue} {User} {Simulator} for {Few}-shot {Data} {Augmentation}}.
\newblock In \emph{Findings of the {Association} for {Computational} {Linguistics}: {EMNLP} 2022}, pages 3788--3799, Abu Dhabi, United Arab Emirates. Association for Computational Linguistics.

\bibitem[{Wu et~al.(2022)Wu, Liao, Qian, and Chua}]{wu_semi-supervised_2022}
Yuxia Wu, Lizi Liao, Xueming Qian, and Tat-Seng Chua. 2022.
\newblock \href {https://doi.org/10.18653/v1/2022.findings-emnlp.462} {Semi-supervised {New} {Slot} {Discovery} with {Incremental} {Clustering}}.
\newblock In \emph{Findings of the {Association} for {Computational} {Linguistics}: {EMNLP} 2022}, pages 6207--6218, Abu Dhabi, United Arab Emirates. Association for Computational Linguistics.

\bibitem[{Xu et~al.(2024)Xu, Guan, Greene, Kechadi et~al.}]{xu2024benchmark}
Cheng Xu, Shuhao Guan, Derek Greene, M~Kechadi, et~al. 2024.
\newblock Benchmark data contamination of large language models: A survey.
\newblock \emph{arXiv preprint arXiv:2406.04244}.

\bibitem[{Yu et~al.(2022)Yu, Wang, Cao, Shafran, Shafey, and Soltau}]{yu_unsupervised_2022}
Dian Yu, Mingqiu Wang, Yuan Cao, Izhak Shafran, Laurent Shafey, and Hagen Soltau. 2022.
\newblock \href {https://doi.org/10.18653/v1/2022.naacl-main.86} {Unsupervised {Slot} {Schema} {Induction} for {Task}-oriented {Dialog}}.
\newblock In \emph{Proceedings of the 2022 {Conference} of the {North} {American} {Chapter} of the {Association} for {Computational} {Linguistics}: {Human} {Language} {Technologies}}, pages 1174--1193, Seattle, United States. Association for Computational Linguistics.

\end{thebibliography}
\bibliographystyle{acl_natbib}

\onecolumn

\appendix

\end{document}